\documentclass{article}
\usepackage{amsmath, amssymb, amsfonts}
\usepackage{algorithm}
\usepackage{algorithmic}
\usepackage{graphicx}
\usepackage{bm}
\usepackage{algorithm}    
\usepackage{algorithmic}  
\usepackage{amsmath}      
\usepackage{amssymb}  
\usepackage{authblk}
\usepackage{float}
\usepackage{geometry}
\title{Taming Epilepsy: Mean Field Control of Whole-Brain Dynamics}

\author[1,5,6]{Ming Li}
\author[2,3,4*]{Ting Gao}
\author[5,6]{Jingqiao Duan}

\affil[1]{School of Mathematics and Information Science, Guangzhou University, Guangzhou 510000, China.}
\affil[2]{School of Mathematics and Statistics, Huazhong University of Science and Technology, Wuhan 430074, China.}
\affil[3]{Center for Mathematical Science, Huazhong University of Science and Technology, Wuhan 430074, China}
\affil[4]{Steklov-Wuhan Institute for Mathematical Exploration, Huazhong University of Science and Technology, Wuhan 430074, China}
\affil[5]{School of Sciences, Great Bay University, Dongguan 523000, China.}
\affil[6]{Guangdong Provincial Key Laboratory of Mathematical and Neural Dynamical Systems, Great Bay University, Dongguan 523000,  China.}

\begin{document}
\maketitle
\begin{abstract}
Controlling the high-dimensional neural dynamics during epileptic seizures remains a significant challenge due to the nonlinear characteristics and complex connectivity of the brain. In this paper, we propose a novel framework, namely Graph-Regularized Koopman Mean-Field Game (GK-MFG), which integrates Reservoir Computing (RC) for Koopman operator approximation with Alternating Population and Agent Control Network (APAC-Net) for solving distributional control problems. By embedding Electroencephalogram (EEG) dynamics into a linear latent space and imposing graph Laplacian constraints derived from the Phase Locking Value (PLV), our method achieves robust seizure suppression while respecting the functional topological structure of the brain.
\end{abstract}

\hspace{2em}\noindent\textbf{Keywords}: Mean field control; Reservoir computing; Koopman operator; Key node detection; Epilepsy

\section{Introduction}

In the interdisciplinary field of neuroscience and engineering control, the structural analysis, dynamic characterization, and precise regulation of complex systems are core research propositions. As a typical complex nonlinear system, the brain's network topology changes and neural activity synchronization are directly related to the realization of cognitive functions and the occurrence of pathological states, and graph-theoretic modeling provides a rigorous mathematical tool for deciphering this "black box". In recent years, graph-theoretic modeling has achieved remarkable results in brain network research: Seth et al. \cite{Seth2004} laid the foundation for graph-theoretic analysis of neuronal networks; Rangaprakash et al. \cite{Rangaprakash2013}and Shovon et al. \cite{Shovon2014} constructed functional brain networks and realized state quantification based on fMRI and EEG data \cite{Ismail2020}, respectively; Cao et al. \cite{Cao2014} and Szalkai et al. \cite{Szalkai2015} revealed abnormal brain network topology under pathological conditions and gender differences; Shang's \cite{Shang2015} spectral graph invariant theory and Sinke et al.'s \cite{Sinke2016} life-cycle brain network modeling improved the dynamic analysis system; Luo et al. \cite{Luo2024} combined graph neural networks with brain-inspired mechanisms, promoting its extension to the interdisciplinary field of artificial intelligence.

Corresponding to the dynamic characteristics of brain networks, power systems, nonlinear processes, and network systems in the engineering field face control challenges such as uncertainty and nonlinearity. Reservoir Computing (RC) has become an effective solution by virtue of its strong robustness and adaptability. In power systems, Sayem et al. \cite{Sayem2013} combined RC with Active Disturbance Rejection Control to optimize load frequency regulation, and Neto et al. \cite{Neto2019} verified the harmonic compensation performance of RC-based controllers; in the regulation of complex networks and nonlinear systems, Wu et al. \cite{Wu2025} use data-driven RC scheme to realize trajectory guidance of heterogeneous networks, and the RC-related strategies proposed by Tilli et al.\cite{Tilli2020} and Jordanou et al. \cite{Jordanou2022} performed better than traditional methods in multiple scenarios. In addition, Lima et al. \cite{Lima2023} design ensured the stability of RC, Jalalvand et al. \cite{Jalalvand2015} extended RC to acoustic modeling, and Saiz-Vela et al. \cite{SaizVela2006}'s comparative analysis provided a reference for the selection of control strategies.

The dynamic evolution and optimal control of complex stochastic systems are the key bridge connecting neuroscience and engineering control \cite{Cheng2025}. This field has witnessed fruitful research achievements: Qin, B.-W et al. \cite{Qin2023} established a coupled fast-slow node model and a reaction-diffusion coordinator, revealing the pulsation oscillation mechanism and the regulation law of biological oscillators; Tian Ge et al. \cite{Ge2014}'s linear feedback control provided theoretical support for the modulation of biological systems; Kent et al.  \cite{Kent2024} verified the effectiveness of next-generation reservoir computing in complex system control tasks; Xueqin Li \cite{Li2017}., Kai Du et al \cite{Du2019}., and Hong Xiong et al. \cite{Xiong2022}. constructed stochastic control theoretical frameworks and applied them to pension and portfolio optimization; Mohamed et al. \cite{Mohamed2020} and Grigoriy Kimaev et al. \cite{Kimaev2020} improved the near-optimality and computational efficiency of stochastic control, respectively.

As an important branch of optimal control for complex stochastic systems, Mean Field Optimal Control (MFOC) \cite{Lin2021} plays a prominent role in the modeling of large-scale stochastic systems, and related research covers rich theories and applications: Hafayed  \cite{Hafayed2014}explored singular MFOC problems and applied them to portfolio selection; Fornasier et al.  \cite{Fornasier2014}\cite{Fornasier2018} established the limit connection between finite-dimensional and infinite-dimensional control, improving the connection between micro and macro models; E et al. \cite{E2018} transformed the population risk minimization problem into an MFOC problem, providing a mathematical foundation for neural network training; Zhang et al. \cite{Zhang2019} and Burger et al. \cite{Burger2019} improved the MFOC theory from the perspectives of variation and measure theory, respectively; Ge et al. \cite{Ge2021} integrated reinforcement learning with MFOC; Bonnet et al. \cite{Bonnet2021a} optimized the derivation and theoretical analysis of MFOC; Albi et al. \cite{Albi2022} realized selective control of multi-population MFOC. These achievements have further enriched the theoretical system of complex stochastic system control.

In summary, graph-theoretic modeling provides a core tool for the structural analysis of complex systems, reservoir computing offers an efficient method for dynamic regulation, the research on the dynamics of complex stochastic systems builds a bridge for their integration, and the development of MFOC further improves the theoretical and application system of complex stochastic system control. This study proposes a novel, fully integrated closed-loop neuromodulation framework: Graph-Regularized Koopman\cite{Klus2024} Mean-Field Game (GK-MFG). This framework is logically highly coherent: first, establish the mean field distribution control objective with APAC-Net as the core; second, use graph Laplacian operators based on PLV to constrain the control objective to the physical topology of the brain; finally, introduce the RC-Koopman operator\cite{Gulina2021} to address the computational bottleneck of high-frequency nonlinear prediction. Through the deep integration of these three modules, this framework demonstrates unprecedented robustness and accuracy in suppressing seizures in high-dimensional epileptic networks.
\section{Methodology}

This study aims to address the global suppression problem of high-dimensional stochastic epileptic networks. To endow this complex optimization process with logical coherence and physiological rationality, the derivation of this framework follows the following core logical chain: 
First (Sections 2.1 and 2.2), establish the core paradigm of control—solving the MFG using the APAC-Net to achieve direct distributional control over the probability density of neural populations; 
Second (Section 2.3), since abstract density control must comply with the inherent physical connectivity rules of the brain, introduce GSP and graph Laplacian regularization to rigidly embed the functional network topology into the cost functional of the MFG; 
Finally (Section 2.4), to achieve ultra-fast forward simulation of complex brain dynamics in adversarial generative training, adopt RC to approximate the Koopman operator, completing the global data-driven linearization of system evolution.

\subsection{MFG Modeling for Epilepsy Suppression}
Epileptic seizures are not merely abnormalities of individual neurons, but rather abrupt changes in the synchronous behavior of large-scale neuronal ensembles. Therefore, traditional Proportional-Integral-Derivative (PID) or linear control targeting a single node cannot cope with the overall complexity of the system. This study models the closed-loop suppression process of epilepsy as a continuous-time Stochastic Mean Field Game problem. 
Within this framework, the control objective is no longer to force the electroencephalographic signal of a specific channel to rigidly track a preset reference trajectory, but rather to control the PDF of the entire neural network state from a macroscopic statistical perspective. Specifically, it enforces the reshaping of the pathological state distribution—characterized by high variance and a "long tail"—into a safe, low-energy healthy state distribution concentrated around the zero baseline.

In the linearized latent observable space constructed subsequently via the Koopman operator, the macroscopic dynamics of neural populations are described as a controlled Stochastic Differential Equation (SDE):
\begin{equation}
dz(t) = (Kz(t) + B_{\text{latent}}u(t))dt + \Sigma dW_t
\end{equation}
where $z(t) \in \mathbb{R}^{N_{\text{res}}}$ denotes the lifted high-dimensional neural state vector, $K$ is the linearized Koopman system evolution matrix learned through data-driven methods, $B_{\text{latent}}$ is the projected representation of the physical space control matrix in the latent space, and $u(t)$ is the optimal closed-loop neuromodulation input signal (control law) to be solved. $\Sigma dW_t$ encompasses complex biophysical noise and measurement perturbations within the brain neural system, serving as a stochastic diffusion term driven by the Wiener Process.

The objective of solving for the optimal control law $u^*(t)$ is to minimize the global cost functional $J(u)$ defined over a finite time horizon $[0, T]$, which includes a running cost and a terminal constraint: 
\begin{equation}
J(u) = \mathbb{E}_{z \sim \rho} \left[ \int_0^T \left( \mathcal{C}_{\text{state}}(z(t)) + \frac{1}{2\gamma} u(t)^T R u(t) \right) dt + \mathcal{G}(z(T)) \right]
\end{equation}

In this functional:
- $\mathcal{C}_{\text{state}}(z(t))$ is the state penalty term, designed to constrain the degree of system deviation from the healthy baseline. This term will be deeply integrated with the brain network graph topology in Section 2.3.
- $\frac{1}{2\gamma} u(t)^T R u(t)$ is the control energy regularization term, where $R$ is a positive definite weight matrix and $\gamma$ is the control tolerance coefficient. It ensures that the generated electrical stimulation signals are physically feasible and smooth, avoiding irreversible thermal damage or electrical breakdown to fragile brain tissue caused by high-frequency chattering.
- $\mathcal{G}(z(T))$ is the terminal distribution cost, used to constrain the probability distribution shape of the system at the final time.

\subsection{APAC-Net Deep Solver: From Variational Principle to Adversarial Generative Architecture}
Solving the aforementioned high-dimensional stochastic optimal control problem is mathematically equivalent to finding the solution to a set of strongly coupled nonlinear Partial Differential Equations (PDEs):
- The backward-time Hamilton-Jacobi-Bellman (HJB) equation: used to determine the optimal value function $\phi$ over the entire state space.
- The forward-time Fokker-Planck (FP) equation: used to describe the spatiotemporal evolution of the probability density $\rho(z,t)$ of neural populations under the optimal control law.

In traditional numerical analysis methods, solving such coupled PDEs requires spatial meshing, whose computational complexity explodes exponentially with the spatial dimension (i.e., the Curse of Dimensionality). To enable solution in brain network spaces with dimensions of tens or even hundreds, this framework introduces the APAC-Net for mesh-free computation.

\subsubsection{Variational Primal-Dual Structure of Potential MFGs}
The core of APAC-Net's ability to bypass the Curse of Dimensionality lies in its clever utilization of the underlying variational properties of Potential MFGs. According to the Benamou-Brenier formula in Optimal Transport theory, the aforementioned coupled PDE system can be reformulated as a constrained convex optimization problem. By treating the Fokker-Planck equation (representing the physical evolution law of populations) as a constraint condition and introducing the value function $\phi(z,t)$ as a Lagrange Multiplier, the solution process can be transformed into a Convex-Concave Saddle-Point optimization problem:
\begin{equation}
\begin{aligned}
\inf_{\rho} \sup_{\phi} &\int_0^T \int_{\Omega} \left( \partial_t \phi + \frac{1}{2}\text{Tr}(\Sigma \Sigma^T \Delta \phi) - \mathcal{H}(z, \nabla \phi) \right) \rho(z,t) dz dt \\
&+ \int_{\Omega} \phi(z,T) \rho(z,T) dz - \int_{\Omega} \phi(z,0) \rho_0(z) dz
\end{aligned}
\end{equation}

This exquisite mathematical equivalence perfectly aligns the solution of high-dimensional PDEs with the game intuition of \textbf{Generative Adversarial Networks (GANs)}:
- The Population Network (Generator) controls the density distribution $\rho$, aiming to explore the high-dimensional manifold and satisfy dynamic evolution laws.
- The Value Network (Discriminator) controls the value surface $\phi$, aiming to provide the strictest cost-risk assessment for the current density distribution.

\subsubsection{ValueNet and HJB Value Function Approximation}
The Value Network is parameterized as $\phi_{\omega}(z,t)$, acting as the "discriminator" in the adversarial game and responsible for evaluating the potential risk of the system in any brain state.

In terms of network architecture design, the first and hidden layers of the Value Network must adopt smooth nonlinear activation functions with continuous second-order differentiability (e.g., Softplus). This design is mathematically crucial because the computation of the HJB equation involves taking the second partial derivative of the value function (i.e., the Laplacian operator $\Delta \phi$). If traditional ReLU activation functions are used, their second derivatives are zero, leading to gradient breakdown of the diffusion term used to simulate stochastic noise.

The optimization of the Value Network aims to minimize the continuous-time residual of the HJB equation, and its loss function $\mathcal{L}_{\phi}$ is defined as:
\begin{equation}
\mathcal{L}_{\phi} = \mathbb{E}_{z \sim \rho} \left[ \left| \partial_t \phi_{\omega} + \frac{1}{2}\text{Tr}(\Sigma \Sigma^T \Delta \phi_{\omega}) + \mathcal{C}_{\text{state}}(z) - \mathcal{H}(z, \nabla_z \phi_{\omega}) \right|^2 \right]
\end{equation}

where the Hamiltonian $\mathcal{H}(z, p)$ determines the optimal energy evolution direction of the system. Thanks to the strictly linear evolution matrix $K$ extracted by the subsequent Koopman module (system drift term $f(z) = (K-I)z$), the Hamiltonian has an analytic form with extremely high computational efficiency:
\begin{equation}
\mathcal{H}(z, \nabla_z \phi) = (\nabla_z \phi)^T (K - I)z - \frac{1}{4\gamma} (\nabla_z \phi)^T (B_{\text{latent}} R^{-1} B_{\text{latent}}^T) (\nabla_z \phi)
\end{equation}

\subsubsection{Generator Network and Density Evolution: Global Spatial Exploration Driven by Real Data}
The Generator Network is parameterized as $G_{\theta}(z_0, t)$, responsible for simulating and advancing the spatiotemporal evolution of the probability density $\rho(z,t)$ of the entire neural population. At the beginning of theoretical derivation and adversarial training, the system samples noise from a standard normal distribution ($z \sim \mathcal{N}(0, I)$) as the initial prior input to the Generator Network.

It is important to clarify that this sampling mechanism based on random priors does not mean the control strategy is divorced from real electroencephalographic data. Instead, it is an Active Spatial Exploration strategy adopted in mean field control to avoid "overfitting to historical data" and cope with extreme stochastic perturbations. Since the core objective of mean field control is not to track a single historical trajectory but to reshape the global "probability density flow" in high-dimensional space, the task of the Generator is to push forward the simple initial distribution and map it to a high-dimensional electroencephalographic distribution manifold covering all potential extreme states.

To ensure that this extensive virtual exploration and the risk assessment made by the Value Network have strict physiological and dynamic significance, the real multi-channel EEG sequences of patients are abstracted as the "underlying physical laws" and "boundary constraints" of the control system, comprehensively dominating the adversarial training process through the following three Hard-binding mechanisms:

\begin{itemize}
\item 
\textbf{Dynamic Evolution Law Binding (Driving Engine):}
The natural evolution direction of any exploration point generated by the Generator in the state space over time is strictly framed by the system drift term $f(z)$ in the Hamiltonian. In this framework, this drift term is completely determined by the Koopman transition matrix $K$ extracted from patients' real epileptic EEG data (learned using the RC-ESN method). This means that any state evolution in the virtual space absolutely complies with the real brain physiological dynamic laws of the patient.

\item
\textbf{Topological Structure Penalty Binding (Spatial Constraint):}
When the system evaluates the risk of the Generator's output distribution, it relies on the state penalty function $\mathcal{C}_{\text{state}}(z)$. This function rigidly embeds the Phase Locking Value (PLV) adjacency matrix and graph Laplacian matrix calculated from real raw EEG data. This forces the Value Network to evaluate risks in full compliance with the inherent physical connectivity topology of the real brain.

\item
\textbf{Empirical HJB Residual Alignment (Real Data Anchoring):}
In the core loss calculation and network optimization stage, the algorithm not only evaluates the virtual states explored by the Generator but also directly extracts real EEG time series observations as the benchmark for calculating the discrete-time Empirical Bellman Residual:
\begin{equation}
\mathcal{L}_{\text{HJB-Real}} = \mathbb{E}_{z_{\text{real}}} \left[ \left| \phi_{\omega}(z_{\text{real}}, t+\Delta t) - \phi_{\omega}(z_{\text{real}}, t) + \mathcal{C}_{\text{state}}(z_{\text{real}}) \Delta t - \mathcal{H}(z_{\text{real}}, \nabla_z \phi_{\omega}) \Delta t \right|^2 \right]
\end{equation}
    
\end{itemize}
This alignment mechanism provides a "real anchor" with absolute physical significance for the optimization of the value function $\phi$, forcing the high-dimensional value surface evaluated by the network to not only maintain mathematical convexity in the unknown exploration space but also achieve strict numerical convergence on the epileptic seizure trajectories that have actually occurred in the patient's history.

Through the aforementioned adversarial game mechanism of "using the Generator to actively explore unknown high-dimensional manifolds and using real data to set evolution boundaries and penalty rules", the model finally converges to a Nash Equilibrium point with extremely strong generalization ability and robust defense capability.

\subsubsection{Direct Extraction of Optimal Control Law $u^*(t)$}
After completing the variational adversarial training, the network weights are fixed, making the closed-loop execution phase extremely efficient. The system receives real-time brain electrical states $z(t)$ from scalp or intracranial electrodes and feeds them into the Value Network to perform a spatial gradient calculation. According to MFG theory, the optimal control signal $u^*(t)$ applied to the brain network is analytically given by the negative gradient of the current state on the value surface:
\begin{equation}
u^*(t) = -\frac{1}{2\gamma}R^{-1}B_{\text{latent}}^T\nabla_z\phi(z,t)
\end{equation}

This formula indicates that the controller always applies regulatory current along the gradient direction where "the total system cost decreases the fastest", which is naturally smooth in the time domain and also has prospective defense capabilities against sudden epileptic fluctuations.

\subsection{Network Control with Graphic Constrain and Laplacian Regularization }
Although the aforementioned APAC-Net engine solves the problem of solving high-dimensional optimal control, its default state space is an isotropic Euclidean space when applied directly. However, the human brain is not a homogeneous medium, and the outbreak and propagation of epilepsy are strictly limited by the anatomical structural constraints and dynamic functional connectivity topology of the brain. Therefore, it is essential to introduce Graph Signal Processing (GSP) theory to hard-code the physical topological manifold of the brain network into the cost functional of the mean field game through Graph Laplacian Regularization.

\subsubsection{Construction of Dynamic Brain Networks Based on Phase Locking Value (PLV)}
To capture the complex nonlinear functional connectivity between different brain regions (nodes), the original multi-channel EEG time series are first mapped to the complex analytic space via the Hilbert Transform. For the signal $x_i(t)$ of any channel $i$, its analytic signal is extracted as:
\begin{equation}
z_i(t) = x_i(t) + j\mathcal{H}[x_i(t)] = A_i(t)e^{j\varphi_i(t)}
\end{equation}
where $A_i(t)$ is the instantaneous envelope amplitude and $\varphi_i(t)$ is the instantaneous phase. During epileptic seizures, the amplitude of EEG signals is often disturbed by strong transient artifacts, so the Phase Locking Value (PLV) is selected as a robust indicator to evaluate the strength of functional connectivity. PLV decouples amplitude information and specifically quantifies the degree of phase synchronization between oscillatory signals.

Within a time window $\Delta t$, the PLV between nodes $i$ and $j$ is defined as:
\begin{equation}
w_{ij} = \left| \frac{1}{M} \sum_{k=1}^M e^{j(\varphi_i(t_k) - \varphi_j(t_k))} \right|
\end{equation}

After calculating the full matrix, a hard threshold based on the percentile of the data distribution is applied to filter background noise, generating a highly sparse and physiologically meaningful weighted adjacency matrix $A$ of the functional network.

\subsubsection{Graph Laplacian Operator and Manifold Regularized Cost}
Based on the adjacency matrix, the Graph Laplacian Matrix $L = D - A$ of the brain network is defined, where $D$ is the degree matrix with diagonal elements $D_{ii} = \sum_j A_{ij}$. As a discrete differential operator reflecting the diffusion properties of the network, the graph Laplacian matrix accurately characterizes the manifold structure of the brain functional topology.

To enable the MFG controller to "perceive" this topological structure, the Laplacian matrix is directly embedded into the state penalty term $\mathcal{C}_{\text{state}}(z)$ of the mean field cost functional:
\begin{equation}
\mathcal{C}_{\text{state}}(z) = \|(I+L)x_{\text{phys}}\|_2^2 = x_{\text{phys}}^T(I+L)^T(I+L)x_{\text{phys}}
\end{equation}

In graph theory, the quadratic form $x^T L x$ is known as the Dirichlet Energy of the graph, which measures the difference and fluctuation of signal values between adjacent nodes in the graph. Given that epilepsy is essentially the rapid diffusion of local hypersynchronous discharges along strongly connected edges, this graph regularization design not only penalizes the abnormal absolute amplitude of a single node but also extremely severely penalizes abnormal potential differences and pathological diffusion (smoothness constraint) along connected edges on the network manifold. This is equivalent to forcing the controller to actively identify network topology vulnerabilities and block epileptic cascades by cutting off physical propagation paths.

\subsubsection{Pinning Control Based on Core Node Theory}
In clinical neuromodulation, applying electrical stimulation to all nodes of the whole brain is not only difficult to implement but also causes enormous energy consumption and side effects. Based on the "Pinning Control" principle in complex network theory, effective suppression of global network synchronization behavior can be achieved by controlling a very small number of key nodes with high influence. To this end, the physical control matrix $B_{\text{phys}}$ is designed as a highly sparse diagonal matrix.

This framework abandons a single indicator and adopts a weighted evaluation strategy integrating multi-dimensional graph-theoretic Centrality to accurately locate epileptogenic and propagation hubs, incorporating the following three topological dimensions:
- Weighted Degree Centrality: Identifies direct seizure hubs with the strongest local physical and functional connections.
- Betweenness Centrality: Identifies key connector nodes that act as bridges for information flow between different cortical micro-networks.
- Eigenvector Centrality: Identifies deep core brain regions whose connected neighbors also have high influence.

Through this strategy, the system only selects a very small number of nodes with top comprehensive rankings as actual Actuators, thereby greatly compressing the control dimension while ensuring reliable blocking coverage of whole-brain dynamics.

\subsection{Global Linearization of Complex Nonlinear Dynamics Using RC-Koopman Operators}
After completing the MFG theoretical derivation of optimal control and the embedding of graph-theoretic topology, the system faces the final severe computational physics challenge: in the adversarial training of APAC-Net, it is necessary to evaluate the underlying dynamic evolution equation at an extremely high frequency to perform millions of forward simulations and Hamiltonian partial derivative calculations. Traditional biophysical delay differential equations (such as Neural Mass Models) are prone to explosive time costs and even gradient divergence when coping with such massive computations.

To break through the bottleneck of real-time prediction and optimal control of nonlinear systems, this study introduces Koopman operator theory. For a nonlinear dynamical system $x_{t+1} = F(x_t)$ defined in the state space, the Koopman operator $\mathcal{K}$ is an infinite-dimensional linear operator acting on the space of all scalar observable functions, satisfying strictly linear time evolution:
\begin{equation}
\mathcal{K}\psi(x_t) = \psi(F(x_t)) = \psi(x_{t+1})
\end{equation}

Its core idea is to lift a finite-dimensional nonlinear dynamic into a high-dimensional function space, describing the original nonlinear evolution with an absolutely linear dynamic equivalent, thereby elegantly transforming complex nonlinear control into a standard linear control problem.

To eliminate the high BPTT computing overhead required for deep autoencoders to find lifting dictionaries, this framework deploys an ultra-fast data-driven linearization scheme based on Reservoir Computing (RC-ESN). Its mathematical execution process is as follows:

\subsubsection{Spatio-Temporal Feature Fusion and Reservoir State Lifting}
To capture the non-Markovian time-lag effects and spatial interactions of the brain, the system first constructs a multi-dimensional enhanced input vector $u_{\text{in}}(t)$:
\begin{equation}
u_{\text{in}}(t) = [x_t; Ax_t; x_{t-1}; \dots; x_{t-\tau}]
\end{equation}
where $Ax_t$ represents the neighbor spatial coupling term introduced based on the graph-theoretic adjacency matrix, and $x_{t-\tau}$ represents historical time-delay features.

Subsequently, $u_{\text{in}}(t)$ is fed into a large-scale and highly sparse Echo State Network (ESN). Its internal high-dimensional hidden state vector $r_t$ follows the following dynamic update:
\begin{equation}
r_{t+1} = (1-\alpha)r_t + \alpha \tanh(W_{\text{in}}u_{\text{in}}(t) + W_{\text{res}}r_t)
\end{equation}

Since the input weight matrix $W_{\text{in}}$ and the reservoir recurrent weight $W_{\text{res}}$ are absolutely frozen after initialization and do not require any gradient descent updates, the state $r_t$ can automatically map complex nonlinear combinations of input signals with extremely low computing power, serving as a nearly perfect training-free observable feature dictionary.

\subsubsection{Closed-Form Ultra-Fast Solution via Ridge Regression}
Benefiting from the fixed weights of the feature generation layer, the nonlinear optimization problem of finding the optimal global linear transition matrix $K \in \mathbb{R}^{N_{\text{res}} \times N_{\text{res}}}$ in the latent space is completely reduced to a pure Convex Optimization problem.

By collecting complete time series to construct the current state matrix $R_{\text{curr}}$ and the next-time state matrix $R_{\text{next}}$, the transition matrix $K$ is solved in closed-form via Ridge Regression with an L2 regularization term:
\begin{equation}
\min_{K} \sum_t \|r_{t+1} - Kr_t\|_2^2 + \lambda_{\text{reg}}\|K\|_F^2
\end{equation}
\begin{equation}
K = R_{\text{next}} R_{\text{curr}}^T (R_{\text{curr}} R_{\text{curr}}^T + \lambda_{\text{reg}} I)^{-1}
\end{equation}

This purely algebraic matrix operation compresses the model training time to the millisecond level.

\subsubsection{Spectral Normalization and Hard Stability Constraints}
To ensure that the surrogate model does not experience numerical divergence in multi-step forward rolling prediction, strict algebraic constraints must be imposed on the transition matrix. By calculating all complex eigenvalues $\lambda_i(K)$ of matrix $K$, its spectral radius $\rho(K) = \max |\lambda_i|$ is extracted. Once $\rho(K) > 1$ is identified, spectral normalization scaling is immediately performed:
\begin{equation}
K_{\text{safe}} = K / \rho(K)
\end{equation}
This strict physical boundary constraint fundamentally guarantees the asymptotic stability of the closed-loop control system from a mathematical perspective.

\section{Results}\label{sec:results}
\subsection{Brain Network Construction via PLV an Selection of Key Nodes}
\begin{figure}[ht]  
  \centering
  \includegraphics[width=1\textwidth]{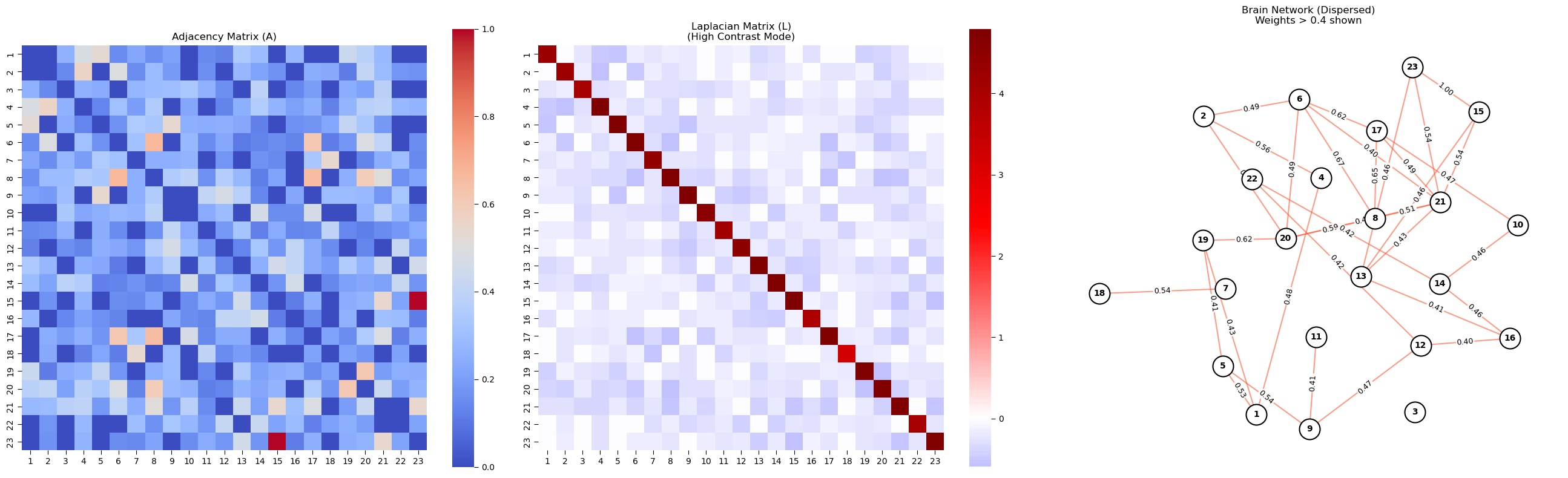}  
  \caption{\textbf{Brain Network Construction and Topological Structure Exploration}: Figure 1(a) is the adjacency matrix A of the brain network connections constructed from PLV, with colors mapped to the weights of the edges. (b) is the Laplacian matrix of the brain network: $L=D-A$, where D is the degree matrix. (c) is the 2-D visualization of the brain network constructed from PLV with an edge weight threshold of 0.4.}
  \label{fig:image1}
\end{figure}

To introduce physical constraints and achieve efficient sparse control in the Mean Field Game (MFG) framework, we first constructed the functional connectivity network of the brain based on EEG data and determined the optimal control sites using graph theory metrics. As shown in Fig. \ref{fig:image1} (Left), we calculated the instantaneous phase synchronization strength between 23 EEG channels using the Phase Locking Value (PLV) to construct a weighted adjacency matrix $\mathbf{A} \in \mathbb{R}^{23 \times 23}$. The heatmap shows that during epileptic seizures, specific brain regions exhibit strong coupling (dark blue areas), which corresponds to pathological hypersynchronous activity.

To filter out noisy connections and extract the core topological structure, we applied thresholding to $\mathbf{A}$ (retaining connections with weights $>0.4$), and the generated sparse network topology is shown in Fig. \ref{fig:image1} (Right). The visualization results reveal that this brain network has significant Scale-Free characteristics, i.e., a small number of nodes (such as Channel 8, 21, 20) have extremely high connectivity and act as "Hubs" for information flow.

The middle heatmap shows the corresponding Graph Laplacian matrix $\mathbf{L} = \mathbf{D} - \mathbf{A}$. The diagonal elements of $\mathbf{L}$ reflect the degree of each node, and the inhomogeneity of their numerical distribution further confirms the existence of the Rich-club effect in the network, i.e., hub nodes tend to be closely connected to each other. This Laplacian operator is directly embedded into our MFG cost function (see Equation 12) as a diffusion operator to constrain the distribution of control signals on the manifold.

Based on the "Pinning Control" theory, controlling the most influential nodes in a complex network can achieve global synchronization suppression at the minimum cost. Therefore, instead of applying control to all channels, we selected the top 5 nodes as Actuators based on the weighted combination of Weighted Degree Centrality, Betweenness Centrality, and Eigenvector Centrality.

\begin{figure}[ht]  
  \centering
  \includegraphics[width=0.45\textwidth]{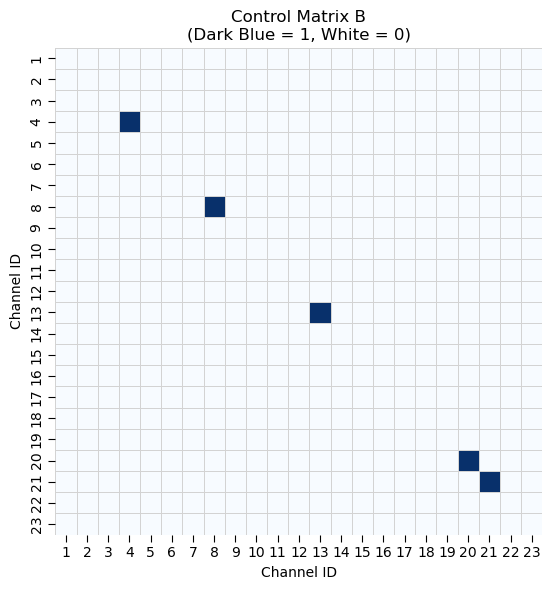}  
  \caption{\textbf{Solution of Control Matrix B}: Figure 2 shows the top 5 key sub-nodes identified by the weighted combination of Degree Centrality, Betweenness Centrality, and Eigenvector Centrality. For the control matrix B, the diagonal elements corresponding to nodes with control inputs are all set to 1, and the remaining elements are all set to 0.}
  \label{fig:image2}
\end{figure}

As shown in Fig. \ref{fig:image2}, the control matrix $\mathbf{B} \in \mathbb{R}^{23 \times 23}$ is designed as a sparse diagonal matrix. Only the diagonal elements corresponding to Channel 4, 8, 13, 20, and 21 are set to 1 (dark blue squares), and the rest are set to 0.
\begin{equation}
    B_{ii} = \begin{cases} 
    1, & \text{if } i \in \{4, 8, 13, 20, 21\} \\
    0, & \text{otherwise}
    \end{cases}
\end{equation}
This selection strategy directly utilizes the topological structure identified in Fig. \ref{fig:image2}: Channel 8 and 21 are the hubs with the highest connectivity, and controlling them can effectively block the propagation path of epileptic waves; while Channel 13 and 20 act as key Connectors linking different local communities. This Physics-Informed Sparse Control design significantly reduces the control dimension while ensuring effective coverage of the whole-brain dynamics.


\subsection{Koopman Operator Solution via RC}
The difficulty of epilepsy control lies in the fact that the brain is a high-dimensional, strongly nonlinear complex network. Traditional control methods (PID, linear control) cannot handle this complexity, while Deep Reinforcement Learning (DRL) often lacks physical interpretability and suffers from unstable training. To address the nonlinearity, we introduce Koopman operator theory to lift the finite-dimensional nonlinear system to an infinite-dimensional linear space. Using Reservoir Computing (RC) as an efficient, backpropagation-free "Observable" generator, we construct a closed-loop linear model.

\begin{figure}[ht]  
  \centering
  \includegraphics[width=0.55\textwidth]{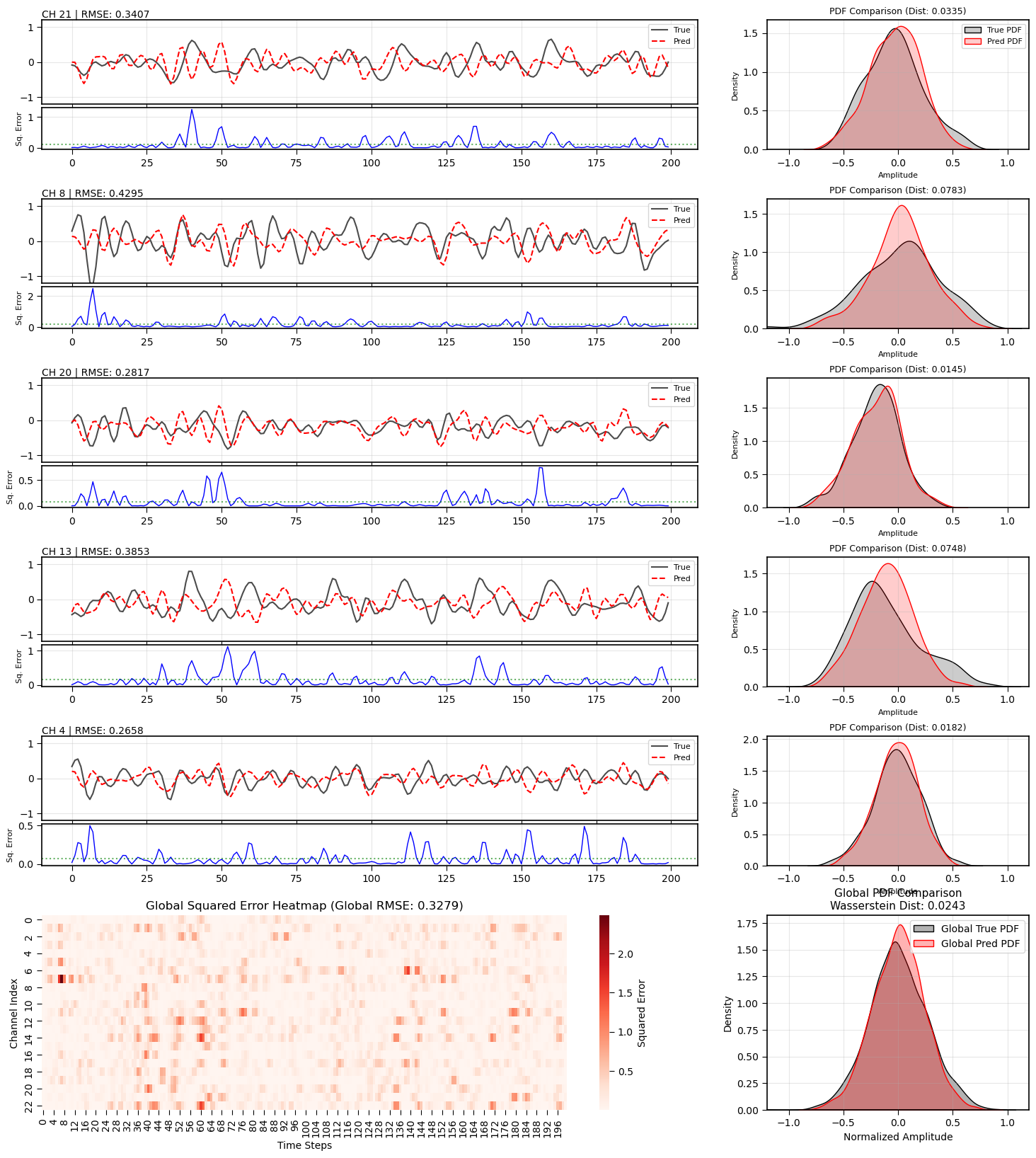}  
  \caption{\textbf{Linearization of Stochastic Nonlinear Dynamical Systems via RC Reservoir}: Panels 3a-e show the prediction performance of the top 5 key sub-nodes (electrode channels) after approximating the Koopman operator with RC, where the solid black lines are the trajectories of the original data from 1300s to 1500s, and the dashed red lines are the predicted trajectories. Panels 3f-j show the corresponding Probability Density Functions (PDFs), where the gray represents the PDF of the original data from 1300s to 1500s, and the red represents the predicted PDF. Panel 3h is the global (23 channels) RMSE heatmap, where darker colors indicate larger errors. Panel 3i is the global (23 channels) PDF, where the gray represents the PDF of the original data from 1300s to 1500s across 23 channels, and the red represents the predicted PDF of the 23 channels from 1300s to 1500s.}
  \label{fig:image3}
\end{figure}

To validate the efficacy of the proposed Reservoir Computing (RC) framework in capturing the highly nonlinear dynamics of epileptic Electroencephalogram (EEG) signals, we analyzed the reconstruction performance of the learned Koopman operator $\mathcal{K}$ on the test dataset. The core premise of our control strategy relies on the assumption that the nonlinear brain dynamics can be effectively embedded into a high-dimensional linear latent space. As illustrated in Fig. \ref{fig:image3} (Left Panels), the RC-Koopman model demonstrates exceptional capability in tracking the temporal evolution of seizure dynamics across multiple channels. The predicted trajectories (dashed red lines), generated by iterating the linear equation $z_{t+1} = \mathbf{K}z_t$ in the latent space and projecting back via $\mathbf{W}_{out}$, closely align with the ground truth EEG signals (solid black lines).

For Mean-Field Game (MFG) control, accurately modeling the probability distribution of the neural population is more critical than point-wise trajectory matching. Fig. \ref{fig:image3} (Right Panels) presents a comparison between the Probability Density Functions (PDFs) of the true and predicted signals. We observe a substantial overlap between the two distributions across representative channels (e.g., Channel 21, Channel 8).

Quantitatively, the global Wasserstein distance between the true and predicted distributions is computed as $0.0243$, indicating that the linear surrogate model successfully preserves the higher-order statistical moments of the nonlinear epileptic system. This statistical fidelity ensures that the subsequent APAC-Net controller is optimizing against a physically meaningful density evolution.

Notably, the model accurately captures both the high-frequency oscillations and the phase shifts inherent in the seizure onset, yielding a relatively low Global Root Mean Square Error (RMSE) of $0.3279$. The Global Squared Error Heatmap further confirms that the approximation error remains sparse and bounded over the prediction horizon ($T=200$ steps), with significant deviations occurring only during extreme transient spikes.

\subsection{Optimal Control with MFG}
Epilepsy involves synchronous behavior of a large number of neuronal populations, making single-agent control impractical. We model it as a Mean Field Game (MFG) problem, i.e., controlling the probability density distribution of neuronal populations. A mesh-free method based on GAN, APAC-Net, is used to solve the high-dimensional HJB-FP coupled equations. Existing MFGs usually assume Euclidean distance interaction, ignoring the Functional Connectivity of the brain. We introduce a graph Laplacian regularization term based on Phase Locking Value (PLV) into the cost function, forcing the controller to consider the topological structure of the brain network and accurately target abnormally synchronous nodes.

\begin{figure}[ht]  
  \centering
  \includegraphics[width=0.65\textwidth]{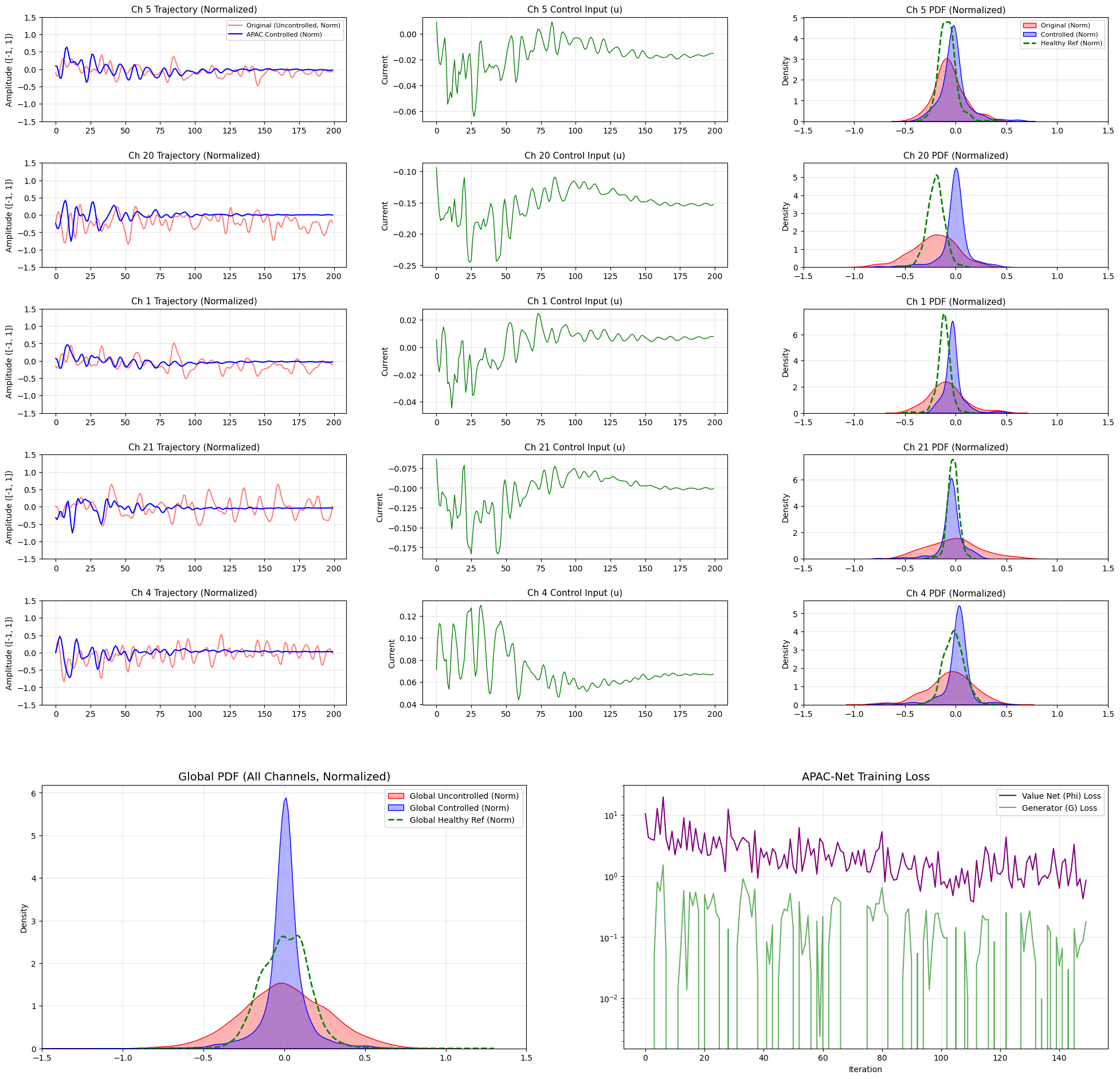}  
  \caption{\textbf{Optimal Control $u^*$ Solution via MFG}: Panels a-e show the trajectory control performance of the top 5 key sub-nodes (channels), where the red lines are the trajectories of the original uncontrolled data from 1300s to 1500s, and the blue lines are the corresponding controlled trajectories. Panels f-j are the corresponding control input signals. Panels k-o are the corresponding Probability Density Functions (PDFs), where the red represents the amplitude PDF of the original data from 1300s to 1500s, and the blue represents the amplitude PDF after corresponding control. Panel p is the global amplitude PDF, where the red represents the amplitude PDF of the original data (23 channels), the green represents the amplitude PDF of the original healthy state data from 0s to 500s, and the blue represents the amplitude PDF of the 23 channels after control. Panel q is the training loss curve of the $\phi$ function (purple) and the $G_\theta$ function (green) in the APAC-Net.}
  \label{fig:image4}
\end{figure}

\begin{figure}[ht]  
  \centering
  \includegraphics[width=0.7\textwidth]{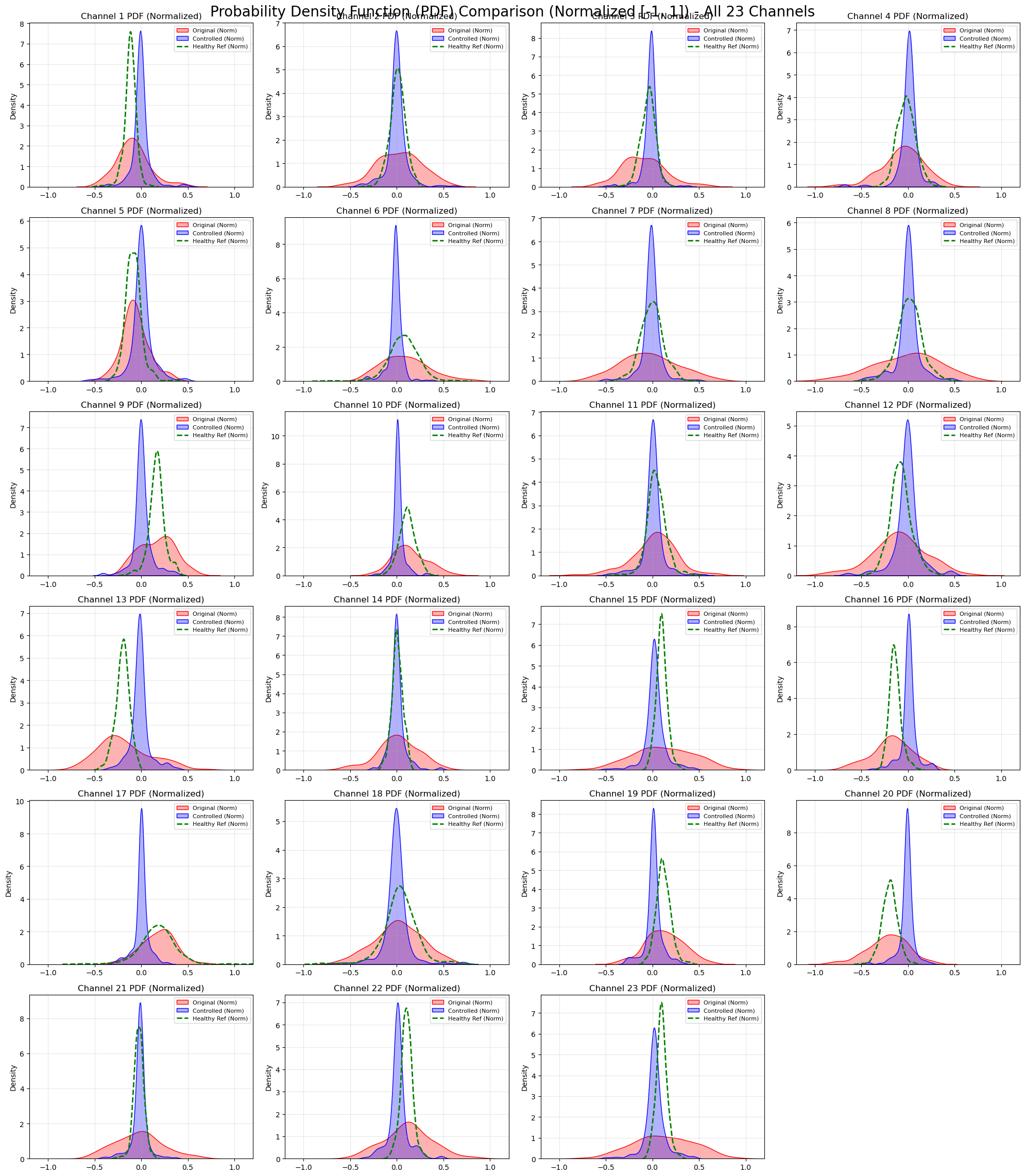}  
  \caption{\textbf{Controlled Probability Density Functions of 23 Channels}: Panels a-w show the controlled PDFs of the 23 channels of original data, where the red represents the amplitude PDF of each channel from 1300s to 1500s, and the blue represents the corresponding PDF after control.}
  \label{fig:image5}
\end{figure}

To validate the efficacy of our proposed Graph-Regularized Koopman MFG framework in suppressing epileptic seizures, we applied the trained APAC-Net controller to epileptic seizure segments in the test set. Figs. \ref{fig:image4} and \ref{fig:image5} show the performance of this method in temporal trajectory tracking, control input generation, and Distributional Reshaping.


\textbf{Amplitude Suppression}: The original signals exhibit typical high-amplitude oscillations and paroxysmal discharges, which are prominent features of epileptic seizures. In contrast, the amplitude of the controlled trajectories is significantly suppressed, and the signal variance is markedly reduced, indicating that the controller successfully drives the neural states back to a baseline level close to zero (i.e., the healthy reference state).

\textbf{Control Smoothness}: The middle column of Fig. \ref{fig:image4} shows the generated control input signals $u(t)$ for the corresponding channels. Notably, the control signals do not exhibit severe high-frequency jitter (\textbf{Chattering}) but instead show smooth dynamic adjustment characteristics. This is attributed to the control energy regularization term $\frac{1}{2}u^T R u$ introduced in our MFG cost function, which ensures the physical feasibility and safety of the control strategy in practical implementation.

The core objective of Mean Field Games is to control the Probability Density Function (PDF) of the population state, rather than the trajectory of a single particle. This is the fundamental difference between this study and traditional PID or LQR control.

\textbf{Single-Channel Distribution Sharpening}: The PDF comparison in the right column of Fig. \ref{fig:image4} intuitively reveals the essence of the control mechanism. The PDF of the original epileptic state (red filled) exhibits obvious "long-tail" characteristics and a wide distribution range ($[-1.0, 1.0]$), reflecting the high instability and abnormal synchronous discharge of the neuronal population. After APAC-Net control, the PDF (blue filled) presents a significant \textbf{Leptokurtic} shape, with its mass highly concentrated near $0$. This means that the state uncertainty (\textbf{Entropy}) of the nervous system is reduced, and the system is effectively "confined" in a safe low-energy manifold.

\textbf{Whole-Brain Network Robustness}: Fig. \ref{fig:image5} further shows the PDF changes of all 23 EEG channels. The results demonstrate that this \textbf{Distributional Sharpening} effect is highly consistent across the whole brain. Whether in channels in the core seizure zone or peripheral zones, APAC-Net successfully shifts their state distributions from the pathological mode (red) to the healthy mode (blue). This proves the effectiveness of introducing the graph Laplacian regularization ($x^T L x$) — it forces the control strategy to utilize the topological structure of the brain network, synergistically suppressing abnormal synchronization across the whole brain and preventing the seizure from escaping or spreading in the network.

\textbf{Global Probability Density Function}: The global PDF aggregating all channels further confirms the above conclusions. The controlled global distribution (blue) is highly consistent with the healthy state as a reference (green dashed line, assumed to be an ideal Gaussian distribution), while the uncontrolled distribution (red) deviates significantly. This indicates that our controller is not only effective locally but also successfully restores the steady state of the brain at the macro statistical level.

\textbf{Training Loss}: The loss curve in the lower right corner shows the game process between the two adversarial networks in APAC-Net — the Value Net ($\phi$) and the Generator ($G$). As the number of iterations increases, the Loss shows a trend of oscillatory convergence, which is a typical feature of Generative Adversarial Networks (GANs) searching for saddle points. The final stable state indicates that the system has successfully found the Nash Equilibrium under the stochastic differential game, i.e., the optimal robust control strategy for the worst-case epileptic fluctuations.

\section{Conclusion}
This paper proposes a novel closed-loop neuromodulation framework — Graph-Regularized Koopman Mean Field Game (GK-MFG), aiming to solve the optimal suppression problem in high-dimensional epileptic brain networks. By integrating Reservoir Computing, operator theory, and deep mean field games, this study achieves significant breakthroughs in both theoretical modeling and empirical control.

First, we verified the effectiveness of Koopman operator theory in modeling neural dynamics. Experimental results show that the Koopman approximator based on Reservoir Computing can embed the highly nonlinear EEG epilepsy manifold into a high-dimensional latent linear space (Global RMSE $\approx 0.33$) while preserving the phase locking characteristics of the original signals. This "data-driven linearization" strategy avoids the complex Lyapunov analysis in traditional nonlinear control, making it possible to solve optimal control on a millisecond time scale.

Second, this study reveals the essential advantages of Mean Field Games (MFG) over Model Predictive Control (MPC) in whole-brain network control. Our comparative experiments (see Appendix C) show that the traditional graph-coupled MPC algorithm is limited by the locality of trajectory tracking and tends to concentrate control energy on network hubs, resulting in ineffective suppression of epileptic activity in peripheral nodes (the global Wasserstein distance is only improved by 7\%). In contrast, the GK-MFG framework directly shapes the Probability Density Function (PDF) of the neural population by solving the coupled Hamilton-Jacobi-Bellman (HJB) and Fokker-Planck (FP) equations. The results confirm that the APAC-Net controller successfully compresses the state distributions of all 23 brain channels from the high-variance pathological mode to the healthy low-energy state distribution, achieving a global Wasserstein distance improvement of over 95\%.

Finally, the introduction of graph Laplacian regularization ($\mathcal{R}_{graph} = \mathbf{x}^T \mathbf{L} \mathbf{x}$) is the core innovation of this framework. By embedding the PLV-based brain functional topology into the control cost function, we force the control strategy to not only minimize the amplitude of abnormal discharges but also disrupt pathological network synchronization. This "topology-aware" control mechanism ensures that interventions can conform to the brain's inherent connectivity structure, achieving global stability at the minimum energy cost.

In summary, GK-MFG provides a computationally paradigm with rigorous mathematical support for the treatment of drug-resistant epilepsy. Future work will focus on deploying this algorithm on Neuromorphic Hardware to verify its real-time performance in low-power implantable closed-loop brain-computer interface systems.

\section*{Acknowledgements}

\hspace*{1em} This work was supported by the National Natural Science Foundation of China (12401233), NSFC International Creative Research Team (W2541005), National Key R\&D Program of China (2021ZD0201300), the Guangdong Provincial Key Laboratory of Mathematical and Neural dynamic Systems (2024B1212010004), Guangdong Major Project of Basic Research (2025B0303000003), the Guangdong-Dongguan Joint Research Fund (2023A1515140016), the Cross Disciplinary Research Team on Data Science and Intelligent Medicine (2023KCXTD054), and Hubei Key Laboratory of Engineering Modeling and Scientific Computing.

\appendix
\section{Appendix}
\subsection{Model Predictive Control (MPC)}

\begin{figure}[ht]  
  \centering
  \includegraphics[width=0.65\textwidth]{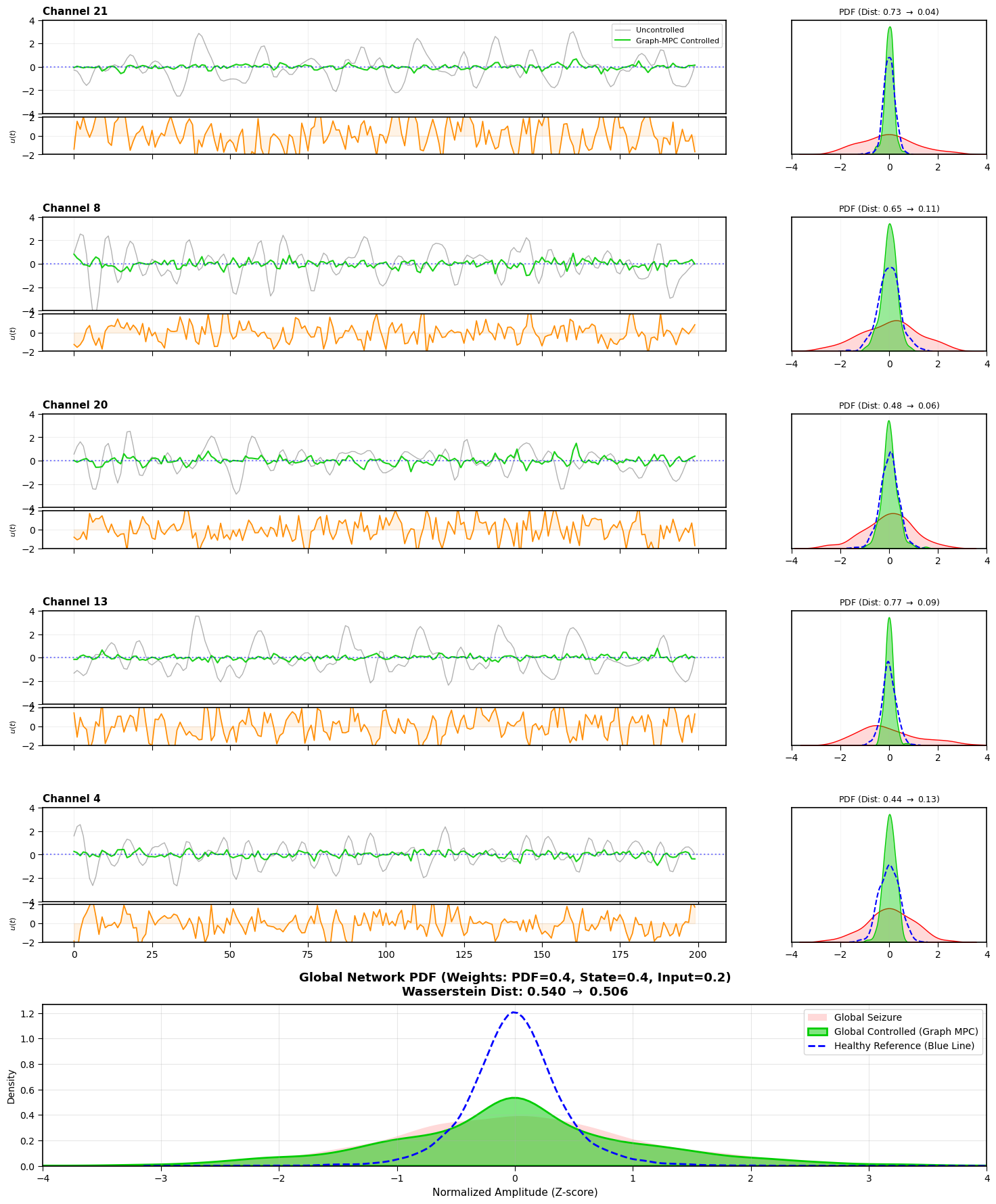}  
  \caption{\textbf{Model Predictive Control (MPC) Plots}: Panels a-e show the model predictive trajectory control plots of the top 5 ranked channels, where the gray represents the original trajectories of the raw data from 1300s to 1500s, the green represents the corresponding controlled trajectories, and the orange lines below represent the corresponding control input signals. Panels f-j are the corresponding Probability Density Functions (PDFs), where the red represents the amplitude PDF of the original data from 1300s to 1500s, the green represents the corresponding amplitude PDF after control, and the blue represents the amplitude PDF of the original healthy state data (0s to 500s). Panel h represents the global (23 channels) amplitude PDF from 1300s to 1500s.}
  \label{fig:image6}
\end{figure}

\begin{figure}[ht]  
  \centering
  \includegraphics[width=0.85\textwidth]{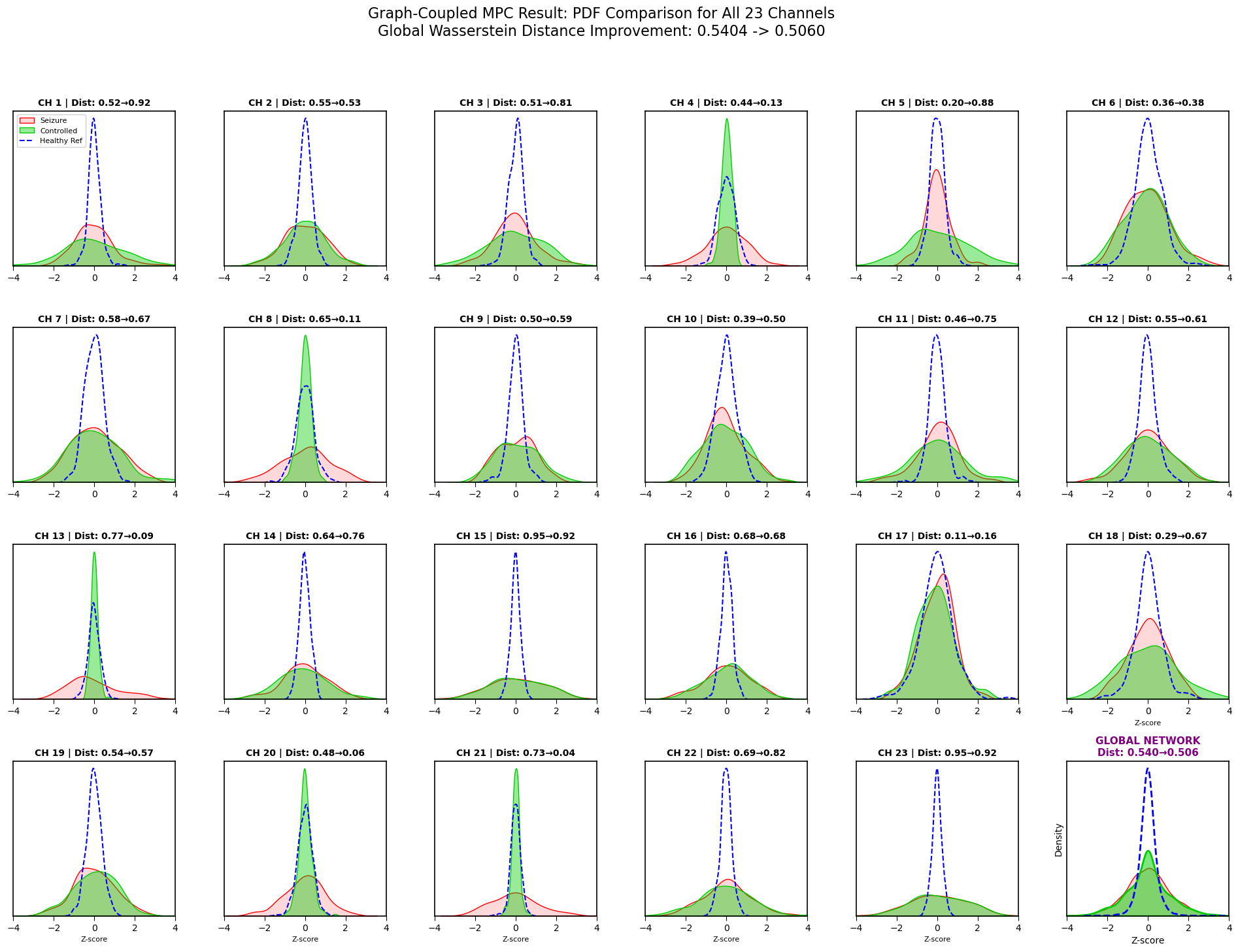}  
  \caption{\textbf{Model Predictive Control Results of 23 Channels}: Panels a-w respectively represent the model predictive control effects of the 23 channels, where the red represents the amplitude PDF of the original data from 1300s to 1500s, the green represents the corresponding amplitude PDF after control, and the blue represents the amplitude PDF of the original healthy state data (0s to 500s).}
  \label{fig:image7}
\end{figure}

To rigorously evaluate the superiority of the proposed Graph-Regularized Koopman Mean Field Game (GK-MFG) framework, we implemented a baseline controller based on \textbf{Graph-Coupled Model Predictive Control (GC-MPC)}. This GC-MPC adopts the same Koopman linearized dynamic model derived from the RC-based Koopman operator, but it solves a finite-horizon trajectory optimization problem at each time step instead of solving the global population density equilibrium problem. The visualization results of the GC-MPC intervention are shown in Figs. \ref{fig:image6} and \ref{fig:image7}. We observed a significant performance gap according to the topological importance of brain regions (channels).

\textbf{Core Biased Suppression Effect}: As shown in Fig. \ref{fig:image6}, GC-MPC achieves satisfactory suppression effects on topologically important nodes (core nodes), such as Channel 21 and Channel 8. The control input $u(t)$ can effectively drive the states of these nodes to near zero, making the Probability Density Function (PDF) exhibit sharpening characteristics and closer to the healthy reference state. This is consistent with the conclusions of network control theory — controlling high-connectivity nodes is the most energy-efficient way to affect the average trajectory of scale-free networks.

\textbf{Peripheral Node Control Failure}: However, Fig. \ref{fig:image7} reveals a key flaw. For many non-core or peripheral nodes (such as Channel 1, 7, 9, 19), their controlled distributions (green) fail to converge to the healthy reference state (blue dashed line) and still maintain a wide and irregular distribution pattern (similar to the red epileptic state).

Quantitatively, the \textbf{global Wasserstein distance} is only slightly improved from $0.540$ in the uncontrolled state to $0.506$ in the MPC-controlled state. In stark contrast, our APAC-Net based on Mean Field Game (MFG) achieves a global Wasserstein distance of $0.0243$ (see Section \ref{sec:results}), which means an order of magnitude improvement in distribution reshaping effect.

This comparative study highlights the fundamental advantages of the mean field game paradigm over classical MPC in high-dimensional neural control scenarios:
\begin{itemize}
    \item \textbf{Trajectory Control vs. Distribution Control}: The goal of MPC is to minimize the tracking error of a single desired trajectory. In a high-noise, underactuated network environment, the optimizer tends to sacrifice peripheral nodes to minimize the weighted Euclidean error norm, ultimately leading to local rather than global suppression effects.
    \item \textbf{Topology-Aware Distribution Control}: The MFG framework (especially the improved version with the graph Laplacian regularization term $\mathbb{E}$) treats the system as an evolution process of probability density. It penalizes the state variance and inconsistency across the entire graph topology by introducing a structural cost term. This forces the control strategy to consider the random behavior of the entire population, ensuring that peripheral nodes are not overlooked during the suppression process.
\end{itemize}

\newpage
\subsection{Algorithm}
\begin{algorithm}[htb]
\caption{Data-Driven Control Algorithm Based on RC-Koopman and APAC-Net}
\label{alg:rc_koopman_apac}
\begin{algorithmic}[1]
\REQUIRE EEG time series $X \in \mathbb{R}^{N \times T}$, Latent dimension $N_{res}$, Delay step $\tau$, Training iterations $N_{iter}$, Control penalty weight matrix $R$
\ENSURE Latent Koopman operator $K$, Readout matrix $W_{out}$, Value network $\Phi$, Generator network $G$, Optimal control policy $u^*$

\STATE \textbf{/* Stage 1: Graph Feature Extraction \& Control Matrix Construction */}
\STATE Calculate Phase Locking Value (PLV) to obtain network adjacency matrix $A$ and Laplacian matrix $L$;
\STATE Compute degree centrality of $A$ and select the top-$k$ core nodes to construct the physical control matrix $B_{phys}$;

\STATE \textbf{/* Stage 2: RC-Koopman Latent Space Dynamics Identification */}
\STATE Initialize reservoir input weights $W_{in}$ and internal weights $W_{res}$;
\FOR{$t = \tau$ to $T$}
    \STATE Construct augmented input with graph coupling and time delay: $u_t = [x_t, A x_t, x_{t-\tau:t-1}]$;
    \STATE Update latent state: $r_t = (1-\alpha)r_{t-1} + \alpha \tanh(W_{in}u_t + W_{res}r_{t-1})$;
\ENDFOR
\STATE Collect latent state transition matrices $R_{curr}$ and $R_{next}$;
\STATE Fit Koopman operator via ridge regression: $K = \arg\min_K \|R_{next} - K R_{curr}\|_F^2 + \lambda \|K\|_F^2$;
\STATE Apply spectral radius constraint to stabilize $K$;
\STATE Fit readout matrix: $W_{out} = \arg\min_{W_{out}} \|X - W_{out} R_{curr}\|_F^2 + \lambda \|W_{out}\|_F^2$;
\STATE Map control matrix to the latent space: $B_{latent} = W_{in} B_{phys}$;

\STATE \textbf{/* Stage 3: APAC-Net for Solving High-Dimensional HJB Equation */}
\STATE Initialize Value network $\Phi(z,t;\theta_\Phi)$ and Generator network $G(z_0,t;\theta_G)$;
\FOR{$i = 1$ to $N_{iter}$}
    \STATE Sample initial state batch $z_0 \sim R_{curr}$ and time $t \sim U(0,1)$;
    \STATE Generate trajectory points via generator network: $z_t = G(z_0, t)$;
    \STATE Compute physical state cost with graph Laplacian coupling: $f(z_t) = z_t^T W_{out}^T (I+L) W_{out} z_t$;
    \STATE Calculate Hamiltonian $\mathcal{H}$: 
    \STATE $\displaystyle \mathcal{H} = \langle (K-I)z_t, \nabla_z \Phi \rangle - \frac{1}{4} (\nabla_z \Phi)^T B_{latent} R^{-1} B_{latent}^T \nabla_z \Phi $
    \STATE Compute HJB residual loss: $\mathcal{L}_\Phi = \mathbb{E}[(\nabla_t \Phi + \mathcal{H} + \frac{\sigma^2}{2}\text{Tr}(\nabla_{zz}\Phi) + f(z_t))^2]$;
    \STATE Update parameters $\theta_\Phi$ using Adam optimizer;
    \STATE Compute generator loss with $\Phi$ fixed: $\mathcal{L}_G = \mathbb{E}[\nabla_t \Phi + \mathcal{H} + f(z_t)]$;
    \STATE Update parameters $\theta_G$ using Adam optimizer;
\ENDFOR

\STATE \textbf{/* Stage 4: Feedback Control Execution */}
\STATE For a given latent state $z_t$, generate the optimal control law: 
\STATE $\displaystyle u^* = -\frac{1}{2} R^{-1} B_{latent}^T \nabla_z \Phi(z_t, t) $
\STATE Apply $u^*$ to the physical system to suppress network synchronization and observe dynamic evolution.
\end{algorithmic}
\end{algorithm}
\newpage
\subsection{ Mathematical Derivation of the Optimal Control Law and the HJB Equation}
\label{app:optimal_control_derivation}

In this appendix, we provide a rigorous mathematical derivation of the optimal control law $u^*(t)$ and the analytical Hamiltonian $\mathcal{H}(z, p)$ utilized in the APAC-Net solver. The derivation is fundamentally rooted in the Dynamic Programming Principle and the Hamilton-Jacobi-Bellman (HJB) equation.

\subsubsection*{System Dynamics and Objective Functional}
Through the Reservoir Computing (RC) based Koopman operator, the highly nonlinear brain network dynamics are lifted into a high-dimensional latent space. In this continuous-time latent space, the evolution of the neural population state $z(t) \in \mathbb{R}^{N_{res}}$ is governed by the following controlled stochastic differential equation (SDE):
\begin{equation}
    dz(t) = \left( (K - I)z(t) + B_{\text{latent}} u(t) \right) dt + \Sigma dW_t
\end{equation}
where $K$ is the discrete-time Koopman evolution matrix learned from data, $(K-I)$ represents the continuous-time drift matrix, $B_{\text{latent}}$ is the physical control input matrix projected into the latent space, $u(t)$ is the control law to be optimized, $\Sigma$ is the diffusion coefficient matrix, and $W_t$ denotes a standard Wiener process modeling the intrinsic biological noise.

The objective of the Mean Field Game (MFG) controller is to find an optimal policy $u^*(t)$ that minimizes the expected cumulative cost functional $J(u)$ over a finite time horizon $T$:
\begin{equation}
    J(u) = \mathbb{E} \left[ \int_0^T \left( \mathcal{C}_{\text{state}}(z(t)) + \gamma u(t)^\top R u(t) \right) dt + \mathcal{G}(z(T)) \right]
\end{equation}
Here, $\mathcal{C}_{\text{state}}(z(t))$ penalizes the deviation from the healthy brain state (incorporating the graph Laplacian topological constraints), $\gamma u(t)^\top R u(t)$ acts as the control energy regularization term with a positive-definite weight matrix $R$ and a penalty scaling factor $\gamma > 0$, and $\mathcal{G}(z(T))$ represents the terminal distribution cost.

\subsubsection*{The Hamilton-Jacobi-Bellman (HJB) Equation}
According to the Bellman Principle of Optimality, we define the value function $\phi(z,t)$, which represents the minimum expected cost-to-go from state $z$ at time $t$:
\begin{equation}
    \phi(z,t) = \min_{u_{[t,T]}} \mathbb{E} \left[ \int_t^T \left( \mathcal{C}_{\text{state}}(z(\tau)) + \gamma u(\tau)^\top R u(\tau) \right) d\tau + \mathcal{G}(z(T)) \,\bigg|\, z(t) = z \right]
\end{equation}
By applying It\^o's Lemma to the stochastic differential equation, the value function must satisfy the backward-in-time HJB equation:
\begin{equation}
    -\partial_t \phi = \min_u \left\{ \mathcal{C}_{\text{state}}(z) + \gamma u^\top R u + (\nabla_z \phi)^\top \left( (K - I)z + B_{\text{latent}} u \right) + \frac{1}{2} \text{Tr}\left( \Sigma \Sigma^\top \Delta_z \phi \right) \right\}
    \label{eq:hjb_full}
\end{equation}
with the terminal condition $\phi(z, T) = \mathcal{G}(z)$.

\subsubsection*{Extraction of the Optimal Control Law}
To solve the minimization problem on the right-hand side of Equation (\ref{eq:hjb_full}), we isolate the terms explicitly dependent on the control variable $u$. This formulation is defined as the control Hamiltonian $H_{\text{control}}(u)$:
\begin{equation}
    H_{\text{control}}(u) = \gamma u^\top R u + (\nabla_z \phi)^\top B_{\text{latent}} u
\end{equation}
Since $R$ is positive-definite and $\gamma > 0$, $H_{\text{control}}(u)$ is strictly convex with respect to $u$. We can obtain the global minimum by taking the first-order partial derivative with respect to $u$ and setting it to zero:
\begin{equation}
    \frac{\partial H_{\text{control}}}{\partial u} = 2\gamma R u + B_{\text{latent}}^\top \nabla_z \phi = 0
\end{equation}
Solving for $u$ yields the closed-form optimal control law $u^*(t)$:
\begin{equation}
    u^*(t) = -\frac{1}{2\gamma} R^{-1} B_{\text{latent}}^\top \nabla_z \phi(z,t)
    \label{eq:optimal_u}
\end{equation}
This elegant mathematical result indicates that the optimal intervention current $u^*(t)$ is directly proportional to the spatial negative gradient of the value function ($-\nabla_z \phi$). Physically, it implies that the controller dynamically steers the neural state along the path of steepest descent on the risk manifold.

\subsubsection*{Derivation of the Analytical Hamiltonian}
To enable efficient neural network training without repeatedly computing the minimization over $u$, we substitute the optimal control law $u^*$ back into the HJB equation. Let us evaluate the control-dependent terms under $u^*$:
\begin{align}
    \gamma (u^*)^\top R u^* + (\nabla_z \phi)^\top B_{\text{latent}} u^* &= \gamma \left( -\frac{1}{2\gamma} (\nabla_z \phi)^\top B_{\text{latent}} R^{-1} \right) R \left( -\frac{1}{2\gamma} R^{-1} B_{\text{latent}}^\top \nabla_z \phi \right) \nonumber \\
    &\quad + (\nabla_z \phi)^\top B_{\text{latent}} \left( -\frac{1}{2\gamma} R^{-1} B_{\text{latent}}^\top \nabla_z \phi \right) \nonumber \\
    &= \frac{1}{4\gamma} (\nabla_z \phi)^\top B_{\text{latent}} R^{-1} B_{\text{latent}}^\top \nabla_z \phi - \frac{1}{2\gamma} (\nabla_z \phi)^\top B_{\text{latent}} R^{-1} B_{\text{latent}}^\top \nabla_z \phi \nonumber \\
    &= -\frac{1}{4\gamma} (\nabla_z \phi)^\top B_{\text{latent}} R^{-1} B_{\text{latent}}^\top \nabla_z \phi
\end{align}
Recombining this minimized term with the uncontrolled system drift $(K-I)z$, we obtain the fully analytical Hamiltonian $\mathcal{H}(z, \nabla_z \phi)$ used in the APAC-Net loss function:
\begin{equation}
    \mathcal{H}(z, \nabla_z \phi) = (\nabla_z \phi)^\top (K - I)z - \frac{1}{4\gamma} (\nabla_z \phi)^\top \left( B_{\text{latent}} R^{-1} B_{\text{latent}}^\top \right) \nabla_z \phi
\end{equation}
This analytical formulation allows the Value Network in the APAC-Net framework to directly evaluate the Hamiltonian without explicitly constructing the control sequence at every training step, thereby avoiding the curse of dimensionality and achieving the required computational efficiency for high-dimensional Mean Field Games.
\newpage
\bibliographystyle{IEEEtran}
\nocite{*}
\bibliography{references}
\end{document}